%% file: coin_2021.tex
\documentclass[runningheads]{llncs}

     \PassOptionsToPackage{numbers, compress}{natbib}

\usepackage[utf8]{inputenc} 
\usepackage[T1]{fontenc}    
\usepackage{hyperref}       
\usepackage{url}            
\usepackage{wrapfig, booktabs}       
\usepackage{amsfonts}       
\usepackage{nicefrac}       
\usepackage{microtype}      
\usepackage{amsmath}
\usepackage{soul}
\usepackage{graphicx}
\usepackage{amssymb}
\usepackage{booktabs}
\usepackage{algorithm}
\usepackage{algorithmic}
\usepackage{rawfonts}
\usepackage{enumerate}
\usepackage{bussproofs}
\usepackage{mathtools}
\usepackage{graphicx}
\usepackage{xspace}
\usepackage{caption}
\usepackage{varwidth}
\usepackage{verbatim}
\def\base{\Pi}
\def\augmented{\Pi'}
\def\Sa{S'}
\def\Sb{S}
\def\Aa{A'}
\def\Ab{A}
\def\Pa{P'}
\def\Pb{P}
\def\Ra{R'}
\def\Rb{R}
\def\La{\lambda'}
\def\Lb{\lambda}
\title{Collaborative Human-Agent Planning for Resilience}

\author{%
  Ronal Singh, Tim Miller \\
  School of Computing and Information Systems\\
  University of Melbourne\\
  Australia \\
  \texttt{\{singhrr,tmiller\}@unimelb.edu.au} \\[2mm]
  Darryn Reid \\
  School of Computing and Information Systems\\
  University of Melbourne\\
  Australia \\
  \texttt{singhrr@unimelb.edu.au} \\
}
\author{Ronal Singh\inst{1} \and  
Tim Miller\inst{1} \and 
Darryn Reid\inst{2}} 
\authorrunning{Presented at COINE 2021}
\institute{University of Melbourne, Australia,
\email{\{singhrr,tmiller\}@unimelb.edu.au}\\ \and
Department of Defence Science, Australia,
\email{Darryn.Reid@dst.defence.gov.au}}

\begin{document}

\maketitle

\begin{abstract}
\input{abstract}

\end{abstract}

\input{intro}
\input{lit}
\input{method}
\input{experiment-design}

\input{results}
\input{conc}
\input{acks}

\bibliographystyle{splncs04}
\bibliography{bib}

\end{document}

%% file: abstract.tex
Intelligent agents powered by AI planning assist people in complex scenarios, such as managing teams of semi-autonomous vehicles. However, AI planning models may be incomplete, leading to plans that do not adequately meet the stated objectives, especially in unpredicted situations. Humans, who are apt at identifying and adapting to unusual situations, may be able to assist planning agents in these situations by encoding their knowledge into a planner at run-time. We investigate whether people can collaborate with agents by providing their knowledge to an agent using linear temporal logic (LTL) at run-time without changing the agent's domain model. We presented 24 participants with baseline plans for situations in which a planner had limitations, and asked the participants for workarounds for these limitations. We encoded these workarounds as LTL constraints. Results show that participants' constraints improved the expected return of the plans by 10\% ($p < 0.05$) relative to baseline plans, demonstrating that human insight can be used in collaborative planning for resilience. However, participants used more declarative than control constraints over time, but declarative constraints produced plans less similar to the expectation of the participants, which could lead to potential trust issues.

%% file: intro.tex
\section{Introduction}
\label{sec:intro}

Intelligent agents assist human operators in complex scenarios, such as task planning to fulfil a set of objectives when supervising teams of semi-autonomous vehicles~\cite{rosenfeld2017intelligent}. The models used by these agents are typically both incorrect and incomplete. They lack details of how to respond in unpredicted and unpredictable situations, such as when they encounter a situation for which they have no training data or have not been modelled explicitly~\cite{reid2018autonomy}.

Humans are far superior at recognising unusual situations and adapting robustly. The ability for a person to provide an intelligent agent with additional context and information at \emph{runtime} that is not part of its model (i.e.\ not encoded by an expert at design-time) would increase the effectiveness and trustworthiness of the agent. Consider an example of a team of autonomous aerial vehicles (UAVs) searching for missing hikers in an area near a small airport. Due to aircrafts leaving and arriving at the airport, certain regions may be `no-fly' zones. However, the underlying task planner has no concept of `no-fly' zones at all. Thus, the question is: how can a human operator input the constraints required to the task planner to avoid the no-fly zones while still achieving its objectives? We call such knowledge \emph{resilience} constraints, as they are constraints that improve the resilience of plans. Following \cite{ibanez2016resilience}, we define \emph{resilience} as the ability to recover from the consequences of an unpredicted or adverse event for a given state of the system. 

We define human-agent collaborative planning for resilient planning in Markov Decision Processes (MDPs). We investigate whether people can recognise limitations of plans, identify how to work around the limitations, and provide resilience constraints to a planner to find plans that adhere to these workarounds. Our longer-term vision is to enable non-AI experts to input this type of knowledge at runtime without modifying the domain theory. The agent can then re-plan to improve the solution using this knowledge.  We hypothesise that constraints represented in LTL are suited to encode resilience constraints, as these logics fit naturally with the way that people describe plan properties. To test this hypothesis, we presented 24 participants with 6 tasks in a scenario involving surveillance by a team of UAVs. In each task, a default plan was presented, and an `intelligence update' provided to the participants outlined additional information not knowable by the planner. Participants were asked to provide resilience constraints without changing the underlying domain model. We encoded participants' responses as LTL constraints and measured the resilience of the resulting plans. Our results showed that participants could provide insight leading to more resilient plans, improving the expected returns by 10\%.

%% file: lit.tex
\section{Related Work}
\label{sec:lit}

This section overviews related literature in both human-agent collaborative planning and encoding domain-control knowledge in planning.

\subsection{Human-Agent Collaborative Planning}

In this paper, we are interested in how to incorporate user input into planners at run time, rather than having to be explicitly required to be coded at design time by expert modellers.

While the idea of machine learning algorithms that learn from user demonstration is a well-studied topic (e.g.~\cite{amershi2014power,rosenfeld2015intelligent}), it is less so in human-agent planning. TRAINS \cite{ferguson1996trains} and TRIPS \cite{ferguson1998trips} systems for \emph{mixed-initiative} planning allowed users to provide input into the planning system while other works include~\cite{ai2004mapgen}. As far as we are aware, Anderson et al.~\cite{anderson2000human} were the first to investigate the combination of human and computational expertise, aimed at avoiding search local optima rather than improving resilience. Hayes et al.~\cite{hayesautonomously} investigate how HTN-like constraints can be learnt from task graphs provided by experts. While it is possible to incorporate this  into our work, we instead use linear temporal logic due to its expressiveness. Icarte et al.~\cite{icarte2018advice} had similar objectives to ours but we are interested in how people represent knowledge (advice) that they have without introducing domain variables, something they did not investigate. De Giacomo et al.~\cite{de2019foundations} use LTL to specify restraining specifications that could assess solutions using features unknown to the agent. They require external sensors to verify the restraining specifications, and they do not require agents to be aware of the constraints, while this is imperative for our agents. Similarly, they do not undertake user studies.

A complementary approach to ours is enabling humans to modify a reward function. Kelley et al.~\cite{kelley2017anapproach} proposed an interface that allows an operator to increase and decrease the agent's reward functions; e.g.\ put a high reward on visiting a desirable area, allowing the operator to create more resilient plans. While they did not perform controlled human behavioural experiments, their approach would be complementary to ours, allowing simple constraints that can be modelled as rewards. For managing large teams of agents, we believe that the approach of using concepts such as LTL constraints would scale better because LTL is a better representation of the way that people describe behaviour \cite{kim2017collaborative}. 

The work from Kim et al.~\cite{kim2017collaborative} is the closest related work to ours, and their experimental setup inspired ours. In their study, participants provided strategies in natural language for solving planning problems. These strategies were encoded as preferences in LTL, and new plans were generated. In five out of the six domains, the collaborative plans were lower cost than the non-collaborative plans. Kim et al.~\cite{kim2017collaborative} argue that  LTL formulae are more suitable for representing how people instruct subordinates and are more succinct than explicitly enumerating the steps of a plan. The difference is the entire aim of the work: Kim et al.~\cite{kim2017collaborative} aim for strategies over entire domains that improve the cost and solving time of plans -- in other words, design-time knowledge from people who understand planning --- while in this work, we aim to increase resilience in individual, unpredicted situations using knowledge from operators.

\subsection{Encoding Domain-Control Knowledge in Planning}
\label{sec:lit:dck}

Domain-independent planners cope with complexity of planning through domain-independent heuristics~\cite{Helmert:FD:JAIR06}. However, domain-independent heuristics are not always well informed; for example, they miss key knowledge when interactions between goals are strong~\cite{geffner2013concise}.

\emph{Domain-specific Control knowledge} (DCK) refers to knowledge for solving a particular problem, and is independent from the domain theory of a problem. It has been shown to increase the scalability of planners~\cite{bacchus:TL:AIJ,nau:shop2:JAIR03}. There are a wide range of formalisms for representing DCK, including macro-actions \cite{coles:online-macros:JAIR207}, abstracted state features for generalised planning \cite{khardon1999learning}, procedural domain-control knowledge \cite{baier-proceduralDCK-icaps07}, hierarchical planning \cite{nau:shop2:JAIR03}, and temporally-extended goals \cite{bacchus:TL:AIJ,coles2011lprpg,camacho2017non}. In this paper, we use temporally-extended goals, specified as linear temporal logic (LTL) formula,
\cite{coles2011lprpg,camacho2017non} to encode human knowledge. LTL-based planning is discussed in Section~\ref{sec:resilient-planning-using-ltl}. As far as we are aware, experiments and applications using such formalisms has been performed only by using domain-control knowledge hand-coded by planning experts at design time, with only a comparatively few experiments looking at how expert users could supply such knowledge \cite{kim2017collaborative}.

%% file: method.tex
\section{Model of Collaborative Planning for Resilience}
\label{sec:method}

In this section, we define our model of resilient planning for MDPs, and extend this to human-agent collaborative planning using temporal constraints. 

\subsection{Resilient Planning}
\emph{Resilience} is the ability to recover from the consequences of an unpredicted or unpredictable event for a given state of the system~\cite{ibanez2016resilience}. In this paper, we are interested in one type of resilient planning: when an automated planner cannot possibly derive a correct solution because it lacks the information and concepts necessary to do so. We assume that the information regarding the unpredicted event is known to a human operator, who can judge the resilience of the current plan. The challenge is that the model is \emph{obsolete}; that is, no longer represents the problem at hand and an automated planner could only derive a correct solution by coincidence because it lacks the information and concepts necessary to represent the knowledge that is available to the operator directly.  This is an important problem in many domains with incomplete models, and there will always be attributes for which automated planning tools cannot plan.

The challenge is that the planner has one model, but the resulting plan is executed in some environment that is unknown to the planner. This is, of course, an ubiquitous challenge in artificial intelligence: models used to make decisions are by definition incomplete simplifications of the world they model. A successful plan may work in one scenario while failing in another, but both correspond to the same scenario in the abstract model space.

To define this concept formally, we first define a conceptual model for planning. We adopt a common model of planning using \emph{Markov Decision Processes} \cite{puterman2014markov}.

\begin{definition}[Markov Decision Process (MDP) \cite{puterman2014markov}]
A Markov Decision Processs (MDP) is a tuple $\Pi = (S, A, P, R, \gamma)$, in which $S$ is a set of states, $A$ is a set of actions, $P(s,a,s')$ is a transition function from $S \times A \rightarrow 2^S$, which defines the probability of action $a$ going to state $s'$ if executed in state $s$, $R(s,a,s')$ is the \emph{reward} received for transitions from executing action $a$ in state $s$ and ending up in state $s'$, and $\gamma$ is the discount factor.
\end{definition}

\begin{definition}[Planning Problem \cite{puterman2014markov}]
A planning problem is a tuple $(\Pi, I, O)$, in which $I \in S$ is the initial state and $O$ is the objective to be achieved. In the simplest case, a \emph{goal-directed MDP} \cite{geffner2013concise}, $O$ is just a set of \emph{goal states}, such as $O \subset S$, but a more common objective is simply to maximise the expected discounted reward \cite{puterman2014markov}. or potentially satisfying preferences over plan trajectories \cite{faruq2018simultaneous}. 

The task is to synthesise a \emph{policy} $\pi : S \rightarrow A$ from states to actions that starts in state $I$ and achieves object $O$.
\end{definition}

\begin{definition}[Resilient Planning]
We define a \emph{resilient planning} problem as a tuple $(\base, \augmented, I, O)$, in which $\base = (\Sb, \Ab, \Pb, \Rb, \Lb)$ is the \emph{base} MDP and $\augmented = (\Sa, \Aa, \Pa, \Ra, \La)$ is the \emph{augmented} MDP. Using $\base$, the automated planner synthesises a policy $\pi : \Sb \rightarrow \Ab$ that achieves the objective $O$ from initial state $I$ over the \emph{augmented} model $\Pi'$. We assume that $\Sb \subseteq \Sa$ and $\Ab \subseteq \Aa$.
\end{definition} 

For a reward-maximising MDP problem, the planner produces a policy $\pi$, with the aim of maximising $\Sigma_{t=0}^{\infty}\gamma'^t R'(s'_t,a'_t,s'_{t+1})$ over all trajectories possible from $\pi$. Note here that the reward function, transition function, etc., are given by the augmented problem $\Pi'$, while the planner only has access to the base problem $\Pi$.

Clearly, the general task of resilient planning is not feasible, as $\base$ and $\augmented$ could model completely different domains; e.g.\ a shipping port vs.\ a manufacturing line. However, given reasonable assumptions about the relationships between $\base$ and $\augmented$, such as $\Sb \subseteq \Sa$ and $\Ab \subseteq \Aa$, the problem becomes more feasible. For example, an optimal policy on $\base$ may still achieve high (but not maximal) rewards on problem $\augmented$ if the transition probabilities and rewards change only slightly. 

\subsection{Human-Agent Collaborative Planning for Resilient Planning}

In our definition of resilient planning, $\Pi$ is the planning model and $\Pi'$ is the actual operation  environment. In \emph{collaborative human-agent planning}, the model $\Pi$ is known to the planner, and the base model $\Pi$ and alternative model $\Pi'$ are both partially known to the human. For example, the human represents the concepts and locations of no-fly zones in their mental model.

\begin{definition}[Collaborative resilient planning]
    \label{defn:collaborative-planning}
We define a collaborative resilient planning problem as a resilient planning problem $(\base, \augmented, I, O)$. The task of the human is to define an augmented initial state $I'$ and objective $O'$ for  the automated planner to synthesis a policy $\pi : \Sb \rightarrow \Ab$ using  $(\Pi, I', O')$,  such that $\pi$ solves the objective $O$ from initial state $I$ under problem $\Pi'$.
\end{definition}

Note the human can modify objectives, but not the \emph{domain} model (transitions). There are two main reasons for only modifying the objectives. First, it treats the planning agent as a black-box, making it easier to design interaction with such agents, as highlighted in \cite{de2019foundations}. Second, we believe that the cognitive demands of modifying action definitions and state spaces would be difficult and error-prone for operators under time pressure, and  require a level of knowledge that unduly restricts usability, while changing objectives may be more feasible~\cite{kim2017collaborative}. 

\subsection{Resilient Planning using Temporal Constraints}
\label{sec:resilient-planning-using-ltl}

A key to solving a collaborative resilient planning problem is how to structure $O'$. We claim that temporal constraints provide a suitable formalism to describe the `shape' of solutions. That is, they propose a structure over the generated plans, rather than just goal states or rewards. In short, planners that support temporal constraints are \emph{domain-configurable planners} in the taxonomy of Nau~\cite{nau2007current}, different from domain-specific planners because they can generalise to other domains, but more flexible than domain-independent planners because they support domain-control knowledge. These claims align with recent works, e.g. \cite{icarte2018advice}.

Temporal constraints encode properties that hold over the whole sequence of states in a solution plan \cite{bacchus:TL:AIJ,coles2011lprpg,camacho2017non}. They are flexible enough to specify liveness constraints (some good thing eventually happens), safety constraints (some bad thing never happens), and fairness constraints (an outcome will happen infinitely often if attempted an infinite number of times). A standard way for expression temporal constraints is \emph{linear temporal logic}.

\begin{definition}[Linear Temporal Logic \cite{pnueli1977}]
Linear temporal logic (LTL) is an extension of propositional logic with modal operators to describe temporal relationships between states on a trajectory. Given a set of atomic propositions $P$, the grammar for LTL is defined as follows:
\[
	\phi ::= p \mid \neg \phi \mid \phi \land \phi \mid \bigcirc \phi \mid \phi \mathcal{U} \phi, 
\]
in which $p \in P$, $\neg$ and $\land$ are logical negation and conjunction respectively, $\bigcirc \phi$ specifies that $\phi$ will hold in the next state, and $\mathcal{U}$ is the `until' operator, for which $\psi \mathcal{U} \phi$ states that $\psi$ will hold until $\phi$ becomes true. New temporal operators can be derived from these basic definitions, such as always achieving a particular state, or achieving a particular state at some point in the future.
\end{definition}

\begin{definition}[LTL Planning]
	An LTL planning problem is a tuple $(\Pi, I, O, \phi)$, in which $\Pi$ describes an MDP, $I$ and $O$ are the initial states and objectives respectively, and $\phi$ is an LTL formula. 
\end{definition}

A policy is valid iff it achieves the objectives and iff the trajectories generated from the policy satisfy the temporal constraint $\phi$. 

There exists several LTL solvers in classical planning \cite{bacchus:TL:AIJ,coles2011lprpg}, non-deterministic planning \cite{camacho2017non}, and planning with MDPs \cite{faruq2018simultaneous}. Other works include~\cite{de2013linear,de2014reasoning,camacho2018finite,camacho2018ltl,de2018automata}. Most of these solvers address the problem by encoding the LTL constraint as an finite state machine (FSM), and  then derive a new planning problem that is the cross product of the FSM and the original planning problem. This product problem can be solved by the underlying solver. The accepting states are reached if a generated trajectory satisfies the LTL constraint. The method of solving is not the focus of this paper.

%% file: experiment-design.tex
\section{Evaluation}
\label{sec:experiment-design}

The goals of our evaluation are: (1) to determine whether participants can identify invalid plans due to the incompleteness of the planning model; and (2) whether participants can improve plan resilience via insights encoded as temporal constraints.

\subsection{Domain}
We simulated an aerial surveillance scenario, based on an application of interest to our industry partner. Participants played the role of an operator controlling multiple unmanned aerial vehicles (UAVs), with the objective of undertaking surveillance on assets and targets (unidentified sea vehicles) in the environment. The UAVs are capable of: (1) navigating between cells specified by coordinates; (2) picking up and dropping pallets; and (3) taking and communicating photos of the targets. The objectives are to take photos of all targets, visit each asset at least once, and when required,  deliver pallets to assets. An intelligent agent assists in achieving the default objectives of surveillance, but the planning model used by the planner makes some assumptions: 1) all targets are unknown; 2) the objectives of taking photos, visiting assets, or delivering pallets can be achieved in any order; 3) the targets and assets move at approximately the same speed; and 4) all UAVs are well resourced to complete their missions. While conceptually the task seems straightforward, the set of possible allocations is combinatorially large, and solving this is difficult without automated assistance for anything other than trivial problems.

\subsection{Protocol}
We asked 24 participants, from different backgrounds such as defence science, and government, to complete six surveillance tasks. Over half had a qualification in computer science. Each task had the same baseline objectives described above, and additional `Operator Preferences', which are ordered preferences that the plans should achieve for this specific task only, but not in general. For example, identifying particular targets before others. The models for each task were specified as a deterministic planning task using PDDL 3.0, and represent the base model $\base$ from Definition~\ref{defn:collaborative-planning}.

\begin{table}[ht]
  \begin{varwidth}[b]{0.45\linewidth}
    \centering
    \small
    \caption{The Six Tasks Completed by Participants.}
    \begin{tabular}{p{1cm}p{4.5cm}}
  	\toprule
  	\textbf{Scen-ario} & \textbf{Description}\\
  	\midrule
  	T1 & Fog in a specific area, limiting visibility to take photos of targets.\\
  	T2 & One target identified as friendly, one identified as hostile, possessing missiles that could take out a drone.\\
  	T3 & One UAV facing heavy headwinds and will take longer than anticipated to reach its target.\\
  	T4 & One UAV out of service and one target is friendly.\\
  	T5 & Two UAVs low on fuel, with different estimates of their likelihood of reaching their targets.\\
  	T6 & Uncertainty about the location of a resource that must be delivered to an asset.\\
  	\bottomrule
  	\vspace{1pt}
  \end{tabular}
    
    \label{tab:tasks}
  \end{varwidth}%
  \hfill
  \begin{minipage}[b]{0.5\linewidth}
    \centering
    \includegraphics[scale=0.3]{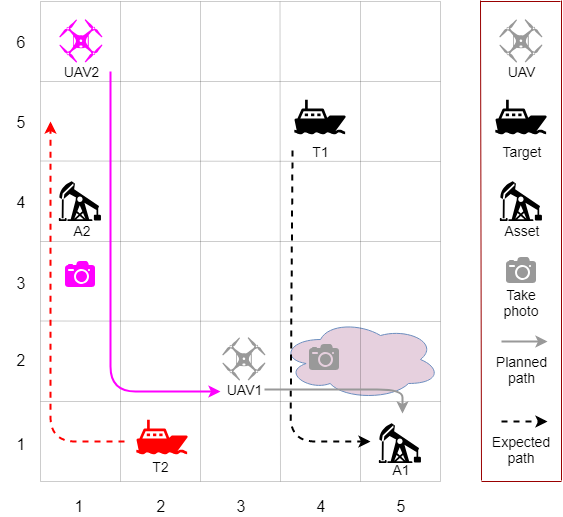}
	
	\begin{flushleft}
		\small
		\textbf{Intelligence update}: Cells (4,2) and (5,2) contain fog. It is unlikely any photo taken in these regions will succeed.
	\end{flushleft}
    \captionof{figure}{Example of one of the tasks used in the study.}
    \label{fig:example-scenario}
  \end{minipage}
\end{table}

\paragraph{Baseline Plan} The participants were presented with a \emph{baseline plan} that achieves the baseline objectives, generated using a state-of-the-art planning that models temporal preferences, LPRPG-P \cite{coles2011lprpg}.  To help the participants visualise this task, the plans were visualised on simple maps that showed the assumed trajectory of targets (if they were non-stationary) and  the proposed plan for the UAVs. One scenario is shown in Figure~\ref{fig:example-scenario}.

\paragraph{Intelligence Updates} Participants received `updates', which are new information derived from intelligence reports that may affect a plan. Importantly, updates contained some information that was not considered in the original model. For example, for the scenario in Figure~\ref{fig:example-scenario}, participants were told that certain coordinates contained heavy fog, and thus taking a picture of the (moving) target in this fog may be unsuccessful. However, the planning model had no concept of `fog', and as such, participants could not simply tell the planner to avoid fog. 
m
\paragraph{Resilience Constraints} For each task, participants were asked to: 1) provide an alternate plan that satisfies the baseline requirements, operator preferences, and the intelligence update; and 2) to write, in natural language, what goals and preferences they would tell the intelligent agent to fulfil so that it would generate the plan provided in item (1). We note that the participant constraints were constraints on the UAV plans and not on the environment, that is, participants could only provide preferences for the UAVs. Table~\ref{tab:tasks} gives an overview of the intelligence updates received for all six scenarios, that is, six participant tasks. The complexity of the aerial surveillance scenario is in the fact that participants had to consider how to state constraints affect the concurrent behaviour of multiple vehicles satisfying the baseline requirements, operator preferences, and the intelligence update. While the participants were required to specify manual plans, they were provided with baseline plans to begin with. This means that they were working with an intelligent planner rather than solving the task from scratch.

\subsection{Procedure}
\label{sec:evaluation-procedure}

The experiment began with a training task using the same domain to ensure that the participants understood the task, and then participants were given six tasks of varying complexity. Participants typically took between 30-40 mins to complete the study although were not given any time limit.

\begin{table}
  \small
  
  \caption{Examples of preference encodings, where UAV1 and UAV2 are unmanned aerial vehicles and T1 is a target.}
  \label{tab:example-encodings}
  \begin{tabular}{l l}
    \toprule
    
    \textbf{Statement} & \textbf{~~~~Encoding} \\
    
    \midrule
    Move UAV1 to cell 4,3 and   & 	\quad(preference p1 (sometime-after \quad(agentloc uav1 v4 v3) \\
    take picture of T1. & \quad (have-photo t1 uav1)))	  \\[2mm] 
    
    UAV1 to take all photos. & 	\quad(forall (?t - target) \\ 
    & \quad (preference p1 \quad (sometime (have-photo ?t uav1))))  \\[2mm] 
    
    D2 to deliver pallet R1 & 	\quad(preference p1 \quad (sometime (carry-pallet r1 d2)))   \\ 
    
    \bottomrule	
    \end{tabular}
\end{table}

The participants' resilience constraints were manually encoded into LTL constraints by the authors, which took an average of 30 minutes for all six tasks. Table~\ref{tab:example-encodings} lists some examples of the participant responses and the encodings using PDDL 3.0. The augmented plans were generated using LPRPG-P \cite{coles2011lprpg} as planning problem $(\base, I', O')$.  We did not set any restrictions on memory and time available to the planner; our problems were solved on a 64-bit machine with 8GB ram and i7 processor. 

\subsection{Assessing Plans}
\label{sec:study-assessingplans}
While the planning tasks were specified as deterministic multi-agent plans, the second model, which we call the \emph{assessment model}, was an MDP, which modelled the rewards for completing tasks and the uncertainties in the environment. Note that we did not provide the assessment model to the participants because we believe that giving all of the underlying MDP model does not reflect reality, and in most cases, this information is not available. 

A baseline deterministic model formed the base of the MDP model, simply using all of the same variables, actions, preconditions, and effects. Then, the following modifications were made. For each goal/preference, a reward of 20 was given, while for each action in a plan, a cost of 1 was applied. For ordering between preferences (e.g.\ each preference A before preference B), a reward of 10 was given if this ordering was preserved. Any tasks containing quantified uncertainties (that is, where participants were given an uncertainty estimate), these uncertainties were used; while for unquantified uncertainties, we used 50\% as the estimate. From this assessment model, the \emph{expected return} of the baseline plan and participants' plans were calculated using a simple custom solver. As well as the baseline plan, participants' plans were compared against an optimal plan, which is a plan that maximises the expected return given the MDP assessment model.

%% file: results.tex
\section{Results}
\label{sec:results}

\subsection{Automated Plans Encoded as LTL}

Table~\ref{tab:automated-results} compares the baseline plans with the plans generated using the resilience constraints. The \emph{Base Plan} is the expected return of the original plan generated without the intelligence update. The \emph{Opt. Plan} is the maximum expected return with respect to the intelligence update, generated using a simple customised solver. The \emph{Auto.\ Plan} shows the average and standard error of the expected return of 24 plans generated from using the automated planner (LPRPG-P) with the participants' constraints encoded as LTL. \emph{Improvement} is the percentage improved between the baseline and manual plan, while \emph{Optimality} is the percentage difference between the optimal and automated plans. The $p$-values are derived using a Mann-Whitney $U$ test (ranked sum test) for unpaired samples.

\begin{table}[ht]
    \centering
    \small
    \caption{Comparing automated plans with baseline and optimal plans. Numbers in brackets represent the standard error.}
    \begin{tabular}{@{~}l@{~}l@{~~}l@{~~}c@{~~~}r@{~}r@{~~}r@{~}r@{~}}
\toprule
                  & \textbf{Base} & \textbf{Opt.} & \textbf{Auto.} & \\
\textbf{Task} & \textbf{Plan}     & \textbf{Plan}    & \textbf{Plan}   &
 \multicolumn{2}{c}{\textbf{Improvement}} & \multicolumn{2}{c}{\textbf{Optimality}}\\
 \midrule
T1	&	72	&	83	&	80.9	 (0.4)	&	12.3\%	&	($p < .01$)	&	-2.6\%	&	($p < .01$)	\\
T2	&	65	&	87	&	70.7	 (1.9)	&	8.8\%	&	($p < .01$)	&	-18.7\%	&	($p < .01$)	\\
T3	&	76	&	92	&	85.8	 (1.2)	&	12.9\%	&	($p < .01$)	&	-6.7\%	&	($p < .01$)	\\
T4	&	63	&	68	&	67.3	 (0.8)	&	6.7\%	&	($p < .01$)	&	-1.1\%	&	($p=.76$)	\\
T5	&	53	&	70	&	58.7	 (0.8)	&	10.8\%	&	($p < .01$)	&	-16.1\%	&	($p < .01$)	\\
T6	&	70	&	80	&	76.9	 (0.3)	&	9.9\%	&	($p < .01$)	&	-3.9\%	&	($p < .01$)	\\
\midrule																
Ave.	&		&		&			&	10.2\%	&		&	-8.2\%	&		\\
\bottomrule
\end{tabular}
    \label{tab:automated-results}

\end{table}

  \begin{table}
    \centering
    \small
    \captionof{table}{Execution times for automated plans. Numbers in brackets represent the standard error.}
    \begin{tabular}{llrr}
    \toprule
    & \multicolumn{3}{c}{\textbf{Time (seconds})}\\
     & \textbf{Base} & \textbf{Opt.} &     {\textbf{Auto.}}\\
    \textbf{Task} & \textbf{Plan} & \textbf{Plan} &     {\textbf{Plan}}\\
\midrule
T1	&	0.32	&	0.40	&	0.7	 (0.1)	\\
T2	&	0.77	&	1.05	&	4.4	 (0.4)	\\
T3	&	0.24	&	460.50	&	317.9	 (40.5)	\\
T4	&	0.79	&	21.00	&	4.9	 (1.7)	\\
T5	&	0.25	&	27.26	&	1.1	 (0.2)	\\
T6	&	1.05	&	2.27	&	2.3	 (0.7)	\\
\midrule								
Ave.	&	0.57	&	85.41	&	55.2	 (21.5)	\\
\bottomrule	
  \end{tabular}
    \label{tab:planner-time-results}
  \end{table}

The results support that LTL encodings are suitable for representing participants' plans. Across all tasks, participants were able to express preferences such that the resulting plans (generated by the planner) were significantly more resilient plans ($p < 0.05$). In only nine instances across all of the 144 tasks did participants preferences not generate an improved plan, and only one had a strictly lower expected return. In this case, it is clear that this was due to a misunderstanding of the intelligence update. On aggregate, the resilient plans were sub-optimal for five of the six tasks ($p<0.05$). For task 4, most of the participants' preferences generated an optimal plan.

The results also demonstrate that participants successfully used the intelligence update to identify problems with the baseline plan and work around this. Over the six tasks, participants were consistently able to specify plans with an improved expected return over the baseline plan. Participants did not always specify the optimal plan. This could be due to the short training session and only six tasks not being enough to learn many subtleties of the domain. Further, participants were not given quantified uncertainties and rewards, so could not have calculated the maximum expected reward optimally. We did not provide the complete assessment model to the participants because in reality, this information is not available. More importantly, our results show that even with incomplete information, participants could still have  an impact.

Table~\ref{tab:planner-time-results} shows the execution times for generating plans. From these results, we can see overall that using resilience constraints (both our own and the participants') take longer to generate than the baseline. This is  because preferences are soft constraints that typically make the problem harder: plans need to satisfy both the baseline goals and the soft constraints, and soft constraints do not prune the search space. The results for task 3 particularly vary significantly between participants, with some plans generated in 1-2 seconds, while others ranged from 300-500 seconds. From our investigations, we believe this is a limitation of the planner rather than the participants' responses. Table~\ref{tab:example-encodings-task3} provides some examples of the encodings and the execution times for some of the responses for task 3. 

\begin{small}
	\begin{table}[!t]
		\centering
		\begin{tabular}{llc}
			\toprule
            
			\textbf{Statement} & \textbf{Encoding} & \textbf{Planning} \\
             &  & \textbf{Time(s)} \\

			\midrule

Assign D1 to take photo of T2   & 	(preference intercept-t1 &	1.11 \\ 
and D2 to take photo of T1 to   & 	\quad (sometime (have-photo t1 d2) )) &	 \\ 
intercept targets at minimum time. & 	(preference intercept-t2 &	 \\ 
 & 	\quad(sometime (have-photo t2 d1))) &	 \\ 
	&	 \\ 
I want to turn D1 off so that D2  & 	(preference not-move-d1 &	460.5 \\ 
takes all photos and visits all assets. & 	\quad(always (agentloc d1 v0 v7)))&	 \\ 
 & 	(preference not-d1-to-t1 &	 \\ 
 & 	\quad(always (not (have-photo t1 d1))))&	 \\ 
 & 	(preference not-d1-to-t2 &	 \\ 
 & 	\quad(always (not (have-photo t2 d1))))&	 \\ 
 & 	&	 \\ 
I want D2 to photos of both targets,  & 	(forall (?t - targets) &	184.8s \\ 
T1 and T2. & 	\quad(preference d2-to-targets &	 \\ 
 & 	\quad\quad(sometime (have-photo ?t d2))))&	 \\ 
		
            \bottomrule	
		\end{tabular}
		\caption{Examples of task 3 encodings and execution times.}
		\label{tab:example-encodings-task3}
	\end{table}
\end{small}	


\subsection{Comparison of Manual and Automated Plans}

Table~\ref{tab:manual-auto-diff} shows the absolute and relative differences of expected return between manual and automated plans. For tasks 3-6, there was no significant difference ($p<0.05$). This is further evidence that LTL formula is a suitable representation for capturing resilience constraints. For task 1, the automated plans had a higher expected return than the manual plans ($p<0.05$). This is the fog scenario from Figure~\ref{fig:example-scenario}. 
\begin{table}
    \centering
    \small
    \caption{Difference between expected return of manual and automated plans. Numbers in brackets represent the standard error.}
    \begin{tabular}{@{~}l@{~~}c@{~~~}c@{~~}r@{~~~}r@{~}r@{~}}
      \toprule
      & Manual & Auto &  \multicolumn{1}{c}{\textbf{Average}} &
      \multicolumn{2}{c}{\textbf{Ave.\ Relative}}\\
      \textbf{Task} & \textbf{Plan} & \textbf{Plan} &
      \multicolumn{1}{c}{\textbf{Diff}} &
      \multicolumn{2}{c}{\textbf{Diff}}\\
      \midrule
T1	&	79.7	 (0.3)	&	80.9	 (0.4)	&	1.2	 (0.5)	&	1.4\%	&	($p < .010$)	\\
T2	&	73.8	 (2.1)	&	70.7	 (1.9)	&	-3.1	 (1.6)	&	-4.8\%	&	($p=.013$)	\\
T3	&	85.4	 (0.9)	&	85.8	 (1.2)	&	0.4	 (0.9)	&	0.2\%	&	($p=.497$)	\\
T4	&	67.5	 (0.4)	&	67.3	 (0.8)	&	-0.3	 (0.4)	&	-0.6\%	&	($p=.192$)	\\
T5	&	58.8	 (0.8)	&	58.7	 (0.8)	&	0.0	 (0.6)	&	-0.2\%	&	($p=.153$)	\\
T6	&	77.0	 (0.2)	&	76.9	 (0.3)	&	0.0	 (0.3)	&	-0.1\%	&	($p=.841$)	\\
\midrule														
Ave.	&			&			&	-0.3	 (0.2)	&	-0.7\%	&		\\
\bottomrule
    \end{tabular}
    \label{tab:manual-auto-diff}
  
\end{table}
In this case, many participants' manual plans moved UAV1 to cell (4,3), then moving three cells to asset A1. However, for most, their resilience constraints stated UAV1 to NOT take a photo in cells (4,2) or (4,3), which allowed the planner to find the more efficient path of taking the photo at cell (4,1), then moving just one cell to asset A1. This demonstrates one advantage of declarative constraints: the planner can search out the best way to achieve the constraint. We believe for more complex scenarios, we would see this more often because specifying a complete control plan would be infeasible. For task 2, we see the inverse: the manual plans have a higher return than the automated plans. The LTL constraints did not control the UAVs in an intended way, however, we believe that had the participants been able to see the generated plans, they would have been able to correct the constraints.

\subsection{Control vs.\ Declarative Preferences}
We analysed the participant responses and classified the resilience constraints as either \textit{control} or \textit{declarative}. Control constraints provide explicit control commands, such as manoeuvring a particular UAV to a particular location, then taking a photo, then going to an asset; while declarative constraints specify a property, such as that the UAV could take a photo of a particular target. There were 32 and 112 \textit{control} and \textit{declarative} types respectively. 

\begin{wrapfigure}{r}{8.5cm}
	\centering
	\includegraphics[scale=0.33]{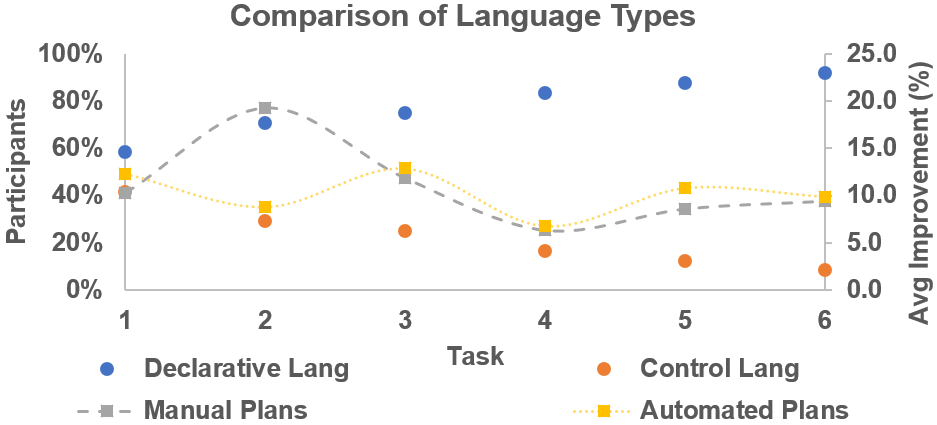}
	\caption{Control vs.\ declarative information and plan similarity}
	\label{fig:plan-similarity}
\end{wrapfigure}

Figure~\ref{fig:plan-similarity} shows that participants used declarative preferences more as they completed more tasks, which could relate to either experience, task complexity, or both. We compared the differences of manual and automated plans, defined as the number of actions present in the manual plan but not in the automated plan. No significant difference was noted in terms of plan differences overall between the preferences types. We compared the plan differences between first and last three tasks, that is, $T1-T3$ (M=5.5, SD=3.2) \& $T4-T6$ (M=6.5, SD=2.9); increased use of declarative preferences leads to significant differences between plans. This is not surprising because with declarative preferences, the planner has more control over the solution. While this is only over six tasks, we expect that this result would hold more generally. Such differences raise potential issues of trust and transparency. That is, such differences, if not understood and accepted by the operator, could decrease the operator's trust in intelligent agents, and ultimately result in disuse. 

%% file: conc.tex
\section{Conclusions}
\label{sec:conc}
We investigated whether people could recognise the limitations of automatically-generated plans and specify constraints on new plans that are more resilient. Twenty-four participants specified constraints for improving baseline plans that had limitations. Our results show that participants' constraints expressed using LTL  improved the expected return of the plans, demonstrating the potential to include human insight into collaborative planning for resilience. In future work, we aim to perform experiments with participants in the loop, enabling them to specify preferences using a restricted language that can be automatically encoded as LTL constraints, and refine the solution iteratively. We will investigate more intuitive and natural methods for eliciting resilience constraints to enable automatic encoding in LTL. We also aim to explore tasks of varying complexity levels in different domains. In terms of presenting the tasks to the participants, we could also different ways, such as randomly presenting the tasks of varying complexity.

%% file: acks.tex
\subsubsection*{Acknowledgements}
\small The research was funded by a Sponsored Research Collaboration grant from the Commonwealth of Australia Defence Science and Technology Group and the Defence Science Institute, an initiative of the State Government of Victoria.